# Robust object extraction from remote sensing data

**Sophie Crommelinck\*, Mila Koeva, Michael Ying Yang and George Vosselman**

Faculty of Geo-Information Science and Earth Observation (ITC), University of Twente, Enschede 7500 AE, The Netherlands; s.crommelinck@utwente.nl (S.C.); m.n.koeva@utwente.nl (M.K); michael.yang@utwente.nl (M.Y.Y.); george.vosselman@utwente.nl (G.V.)
\*Correspondence: s.crommelinck@utwente.nl; Tel.: +31-53-489-5524

**Abstract:** The extraction of object outlines has been a research topic during the last decades. In spite of advances in photogrammetry, remote sensing and computer vision, this task remains challenging due to object and data complexity. The development of object extraction approaches is promoted through publically available benchmark datasets and evaluation frameworks. Many aspects of performance evaluation have already been studied. This study collects the best practices from literature, puts the various aspects in one evaluation framework, and demonstrates its usefulness to a case study on mapping object outlines. The evaluation framework includes five dimensions: the robustness to changes in resolution, input, location, parameters, and application. Examples for investigating these dimensions are provided, as well as accuracy measures for their qualitative analysis. The measures consist of time efficiency and a procedure for line-based accuracy assessment regarding quantitative completeness and spatial correctness. The delineation approach to which the evaluation framework is applied, was previously introduced and is substantially improved in this study.

**Keywords:** accuracy assessment; method evaluation; object delineation; building extraction; boundary extraction; cadastral mapping

## 1. Introduction

The extraction of object outlines from remote sensing data is of interest in numerous application fields such as topographic and cadastral mapping, as well as in urban planning. Delineating objects such as roads, buildings, walls or rivers has been investigated in photogrammetry, remote sensing and computer vision [1,2]. Advances have been achieved through methods of template-matching, mathematical morphology, active contours, Object-Based Image Analysis (OBIA), (un)supervised classification and machine learning [1]. Despite these advances, the problem of automatically extracting object outlines from remote sensing data has not been solved.

One object type often investigated is a building. A large number of studies demonstrates the need and challenge to delineate this object: buildings are extracted from different data sources, such as rasters [3-5], 3D point clouds [6-8], a combination of 2D and 3D data [9-13] or existing maps [14-16]. They are extracted from data captured from different platforms, such as Unmanned Aerial Vehicles (UAV) [17], aircrafts [18-20], or satellites [21-23]. Extraction approaches in such studies often consist of a workflow comprising image segmentation, line extraction and contour generation [24]. The variability of methods and workflows reflects the problem's complexity, consisting of extracting different objects from various data sources. Even for one object type, such as a building, numerous characteristics exist that hinder the compilation of a generic model and thus the development of a generic method. To promote method development, recent work states a need to generalize the delineation problem and to develop reusable evaluation frameworks [25,26].

Existing evaluation approaches calculate accuracy measures per pixel, per object, or on a combination of both [27]. The accuracy measures are obtained by calculating the distance or overlap between the obtained result and the reference data. This allows to quantify the pixels or objects being True Positive (TP), True Negative (TN), False Positive (FP) or False Negative (FN) and to derive accuracy measures [1,9,27-32]. These measures are widely used and considered, e.g., for a benchmark dataset of aerial data for urban object classification and 3D building reconstruction [33], for one of satellite imagery for road extraction, building extraction and land cover classification [34], and for one of aerial imagery for building extraction [35]. The same holds true for suggestions on accuracy assessment by public institutions such as Geoscience Australia [36]. According to Foody, the confusion matrix, from which accuracy measures can be derived, often lies at the core of the accuracy assessment without questioning its suitability and sufficiency [37]. Further aspects are seldom investigated in a comprehensive and structured way. Möller et al. similarly observe a deficit in commonly accepted protocols for accuracy measures [38].

This study structures evaluation dimensions in a comprehensive evaluation framework that investigates an approach's robustness to changes in resolution, input, location, parameters, and application. Each of these dimensions is evaluated with the same assessment measures consisting of quantitative completeness, spatial correctness, and time efficiency. The framework is demonstrated on a boundary delineation approach [39]. This approach supports the delineation of objects by automatically retrieving information from UAV-based RGB and Digital Surface Model (DSM) data that is used to guide an interactive boundary delineation. The boundary delineation approach that is extended in this study has been designed to support the delineation of cadastral parcel boundaries from UAV data. Cadastral boundaries capture the extent of land ownership and are often demarcated by visible objects [40-42]. Example objects used for demarcation are hedges, walls, roads, footpaths, water drainages, pavement, ditches, fences, as well as buildings [24]. In cadastral mapping, which is used as an example application in this study, building footprints and outlines are often captured in addition to walls and fences circumscribing cadastral boundaries (Figure 1). In the remainder of this study, the term building outline is synonymous with roof outline, not with building footprint and the terms of extraction, detection, and delineation of outlines are used synonymously.

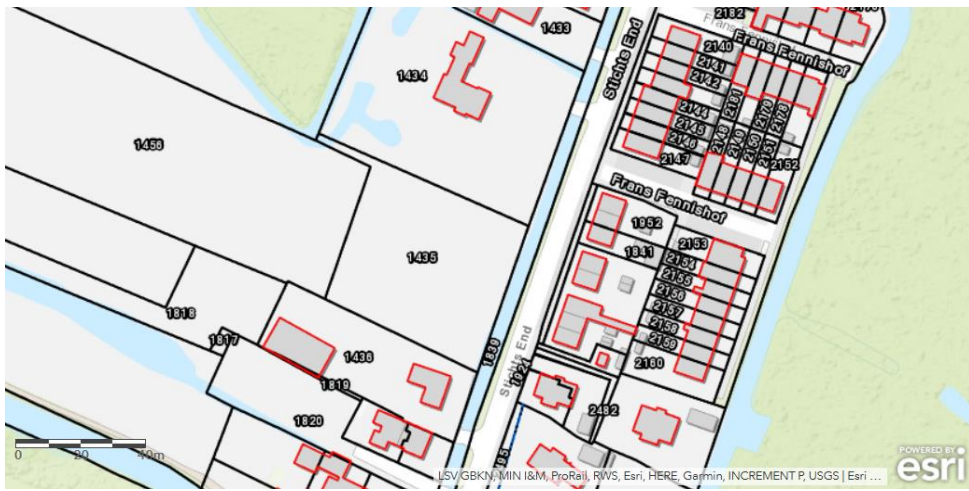

**Figure 1.** Cadastral map from the Netherlands capturing building outlines (red) and parcel boundaries (black) overlaid on a topographic base map [43].

## 2. Materials and Methods

### 2.1. Optical Sensor Data

An urban and a peri-urban area in Rwanda were selected for this study (Table 1, Figure 2). The data were captured with indirect georeferencing, i.e., Ground Control Points (GCPs) were distributed within the field and measured with a Global Navigation Satellite System (GNSS). All flights were carried out with a UAV with 80% longitudinal and 80% transversal overlap. For each mission, the flight height was set to 100 m above the surface. RGB orthomosaics and DSMs of 5 cm Ground Sample Distance (GSD) were generated with Pix4DMapper. Details on the data capture are provided in [44]. Three tiles of 250 x 250 m of both areas were cropped out of the RGB orthomosaics and the DSMs.

| Location | UAV Model | UAV Type | UAV Sensor | Area [ha] | GSD [cm] | Capture |
|----------|-----------|----------|------------|-----------|----------|---------|
| Busogo, Rwanda | FireFLY6 BirdsEyeView | Hybrid | SONY A6000 | 94 | 5 | January 2018 |
| Mukingo, Rwanda | InspirePro DJI | Rotary-wing | Zenmuse X5S | 50 | 5 | January 2018 |

**Table 1.** Specifications of UAV-captured optical sensor data.

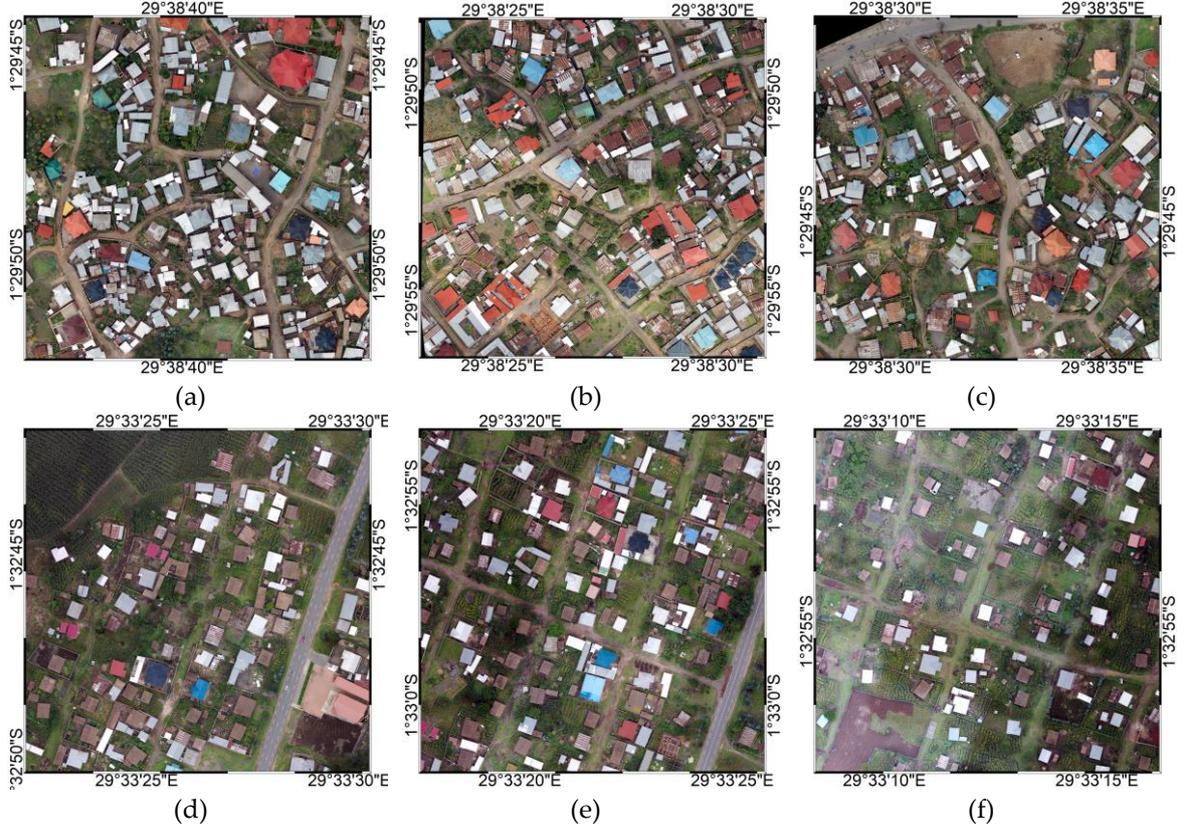

**Figure 2.** UAV data tiles of 250 x 250 m from Busogo **(a-c)** and Mukingo **(d-f)** in Rwanda.

## 2.2. Boundary Delineation Approach

The boundary delineation approach also referred to later as the initial approach is based on [39]. It supports the delineation of boundaries by automatically retrieving information from RGB and DSM data that is then used to guide an interactive delineation. It consists of three parts (Figure 3): (i) image segmentation, (ii) boundary classification and (iii) interactive delineation. The sustainable use of the approach is ensured and facilitated by being implemented in an open-source Docker container. The source code is publically available [45].

(i) **Image segmentation** delivers closed contours capturing the outlines of visible objects in the image. The workflow described in [39] proposes to use Globalized Probability of Boundary contour detection (gPb) [46] and Simple Linear Iterative Clustering superpixels (SLIC) [47]. We now propose to use an extended version of gPb developed by the same authors: Multiresolution Combinatorial Grouping (MCG) [48]. This allows combining the previous steps into one method while increasing spatial accuracy compared to using gPb and decreasing over-segmentation compared to using SLIC.

(ii) **Boundary classification** requires labeling the outline contours from (i) into 'boundary' and 'not boundary' to generate training data. A set of features is calculated per line capturing its geometry (i.e., length, number of vertices, azimuth, sinuosity) and its spatial context (i.e., gradients of RGB and DSM underlying the line). A description of each feature can be found in Table A1 in the appendix. These features together with the labels are used to train a Random Forest (RF) classifier [49]. The trained classifier predicts boundary likelihoods for unseen testing data for which the same features have been calculated. From the tiles in Figure 2, a/d are used for training, b/c/e/f for testing. An open-source RF implementation [50] is used.

(iii) **Interactive delineation** allows a user to start the actual delineation process: the RGB orthomosaic is displayed to the user, who is asked to select nodes to be connected to a boundary. A least-cost-path algorithm searches for the lines from (a) that connect the user-selected nodes taking into account the boundary likelihood from (b). The line is suggested to the user with the options to edit, save or delete. We implemented (c) as publically available plugin [51] for the open-source geographic information system QGIS [52].

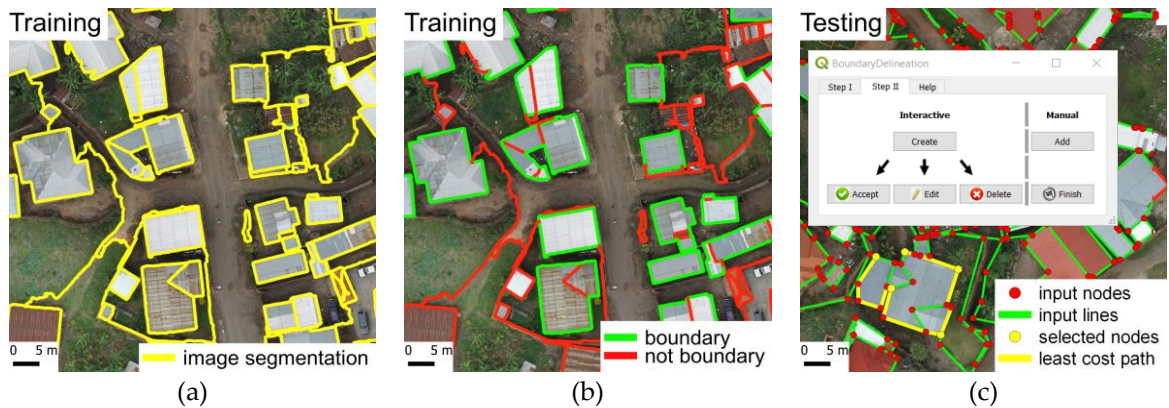

**Figure 3.** Boundary delineation approach: **(a)** MCG **image segmentation**. **(b) boundary classification** that requires line labeling into 'boundary' and 'not boundary' for training. The labeled lines are used together with line-based features to train a Random Forest classifier that generates boundary likelihoods for testing. **(c) interactive delineation** guided by a QGIS plugin that creates a least-cost-path between user-selected nodes along simplified lines from (a) with highest boundary likelihoods generated in (b).

### 2.3. Evaluation Framework

The evaluation framework provides an approach to assess the performance of object delineation methods. A method is investigated in five dimensions: robustness to changes in (I) resolution, (II) input, (III) location, (IV) parameters, and (V) application. All dimensions are evaluated with the same assessment measures consisting of (i) quantitative completeness, (ii) spatial correctness, and (iii) time efficiency.

(I) **Resolution Robustness:** *Can the approach be applied to data of a different resolution?*
Resolution robustness investigates whether an approach developed and tested on data of a specific resolution is applicable to data of a different resolution. One location can be captured from different platforms (e.g., UAV, aircraft and satellite), as well as in different resolutions (e.g., 1-100 cm). Different resolutions can be achieved not only by altering the platform, but also by altering the sensor or the flying height. High-resolution data from satellites often ranges from 30-100 cm, while that from aerial surveys ranges from 2-90 cm. Nowadays, a resolution of 30 cm is the industry standard for most archived large image databases such as Google Earth. This resolution is achievable from aircraft and satellite platforms. For these reasons, a resolution of 30 cm is chosen in this study: the boundary delineation approach that has been tested on UAV data of 5 cm is thus tested on UAV data of 30 cm down-sampled with nearest neighbor resampling. The down-sampled data does not fully mimic aerial or satellite data, as down-sampled data mostly holds more information than data directly captured in that resolution. Furthermore, the data are influenced by the down-sampling approach that determines which and how pixels are aggregated. An ideal evaluation would require cloudless aerial or satellite data of the same timestamp and location.

(II) **Input Robustness:** *Can the input data required for the approach be altered?*
Input robustness investigates whether the amount of required input data or input information can be extended or reduced. One location can be covered by data containing different information (e.g., RGB, DSM, topographic maps, existing delineation maps), whose integration or reduction can influence the results. In this study, the additional value of DSMs is investigated, by applying the initial approach on the 5 cm resolution data once with and once without integrating DSM features during the boundary classification step.

(III) **Location Robustness:** *Can the approach be applied in a different location?*
Location robustness investigates whether an approach developed and tested in one location can also be applied in a different one. The different location can contain new objects to be delineated (e.g., roads, fences, trees) in a new context (e.g., urban, peri-urban, rural). In this study, the initial approach requires a training phase of the RF classifier that considers line features calculated from the corresponding RGB and DSM data. To investigate the location robustness, the boundary classifier trained on one location (Mukingo), is used for the subsequent interactive delineation in a different location (Busogo). Moreover, the approach has been successfully applied to delineate roads in UAV data from Germany [39] and will now be tested to delineate buildings, walls, and fences in UAV data from Rwanda.

(IV) **Parameter Robustness:** *Does the approach depend on many or sensitive parameters?*
Parameter robustness investigates whether the approach depends on many parameters and to what extent changing these parameters influences the result's quality. In this study, parameter robustness is investigated by analyzing the number of parameters in the boundary delineation approach, and the sensitivity of delineation results to small changes in parameter values.

(V) **Application Robustness:** *Can the approach be applied to delineate different objects?*
Application robustness investigates whether an approach suggested for delineating of a specific object can be used on different objects as well. In this study, the applicability of the approach was first investigated for building outlining and now for cadastral mapping: the boundary delineation approach is applied to delineate visible cadastral boundaries demarcated through walls and fences. The official cadastral reference data for the study areas have been digitized in 2009 on aircraft orthomosaics of 25 cm GSD and is thus outdated and of low spatial accuracy. Two Rwandan surveyors with delineation expertise in surveying generated new cadastral reference data for the accuracy assessment of this study. To do so, they applied indirect surveying: parcel boundaries were digitized on the UAV orthomosaic based on local knowledge of boundary locations and characteristics.

### 2.4. Assessment Measures

(i) **Quantitative Completeness:** *How many objects can be delineated entirely?*
Quantitative completeness investigates the percentage of objects that can be entirely delineated by the approach. As reference data, the manually digitized outlines of all undamaged buildings having an area of >25 m² are used to investigate resolution, input and location robustness. As for the application robustness focusing on cadastral mapping, manually digitized visible parcels are used as reference data. Subsequent to the interactive delineation, the number of objects that are either directly acceptable without further editing, require editing, or require manual delineation are summated.

(ii) **Spatial Correctness:** *How correct is the delineation in a spatial sense?*
Spatial correctness investigates to what extent successfully delineated objects coincide with the reference data in a spatial sense. This is done by first buffering the reference data. The buffer size should be chosen in accordance with the required accuracy of the object delineation. For cadastral mapping, the statutorily requested accuracy depends on the needs and nature of the area being surveyed. The measuring approach and the accepted result can vary: the accepted accuracy amounts to 6 cm in the Netherlands, 3-7 cm in Germany and Switzerland, 5-10 cm in Malaysia, 10-30 cm in Poland [53] and 240 cm for rural areas proposed by the IAOO [54]. Subsequently, the percentage of the delineation line lying inside and outside of the reference buffer is calculated. This can be done either vector- or raster-based. For a raster-based approach, as done in this study, the delineation lines are rasterized to the 5 cm GSD of the input data. Only those delineation lines from (i) that were directly acceptable without further editing are used. For the reference lines, a buffer size of 30 cm radius is chosen before rasterizing and overlaying both the delineation and the reference data (Figure 4). The overlay allows labeling pixels as TP, where the delineation line falls inside the reference buffer and as FP, where the delineation line falls outside the reference buffer. The sum of pixels with the same label is summarized in a confusion matrix. From the confusion matrix, the error of commission (Figure 4a) and the correctness (Figure 4b) are calculated in range [0; 100]. The error of commission captures the percentage of pixels falling outside of the reference buffer, the correctness those that fall inside of it.

These measures are based on [55], described as commonly reported measures in more recent publications [30,56], and similarly used to evaluate results of a ISPRS benchmark dataset [2]. We implemented the described procedure for line-based accuracy assessment consisting of buffering, rasterizing, overlaying, calculating and plotting the confusion matrix, in a publically available plugin (Figure 4c) [57] for the geographic information system QGIS [52].

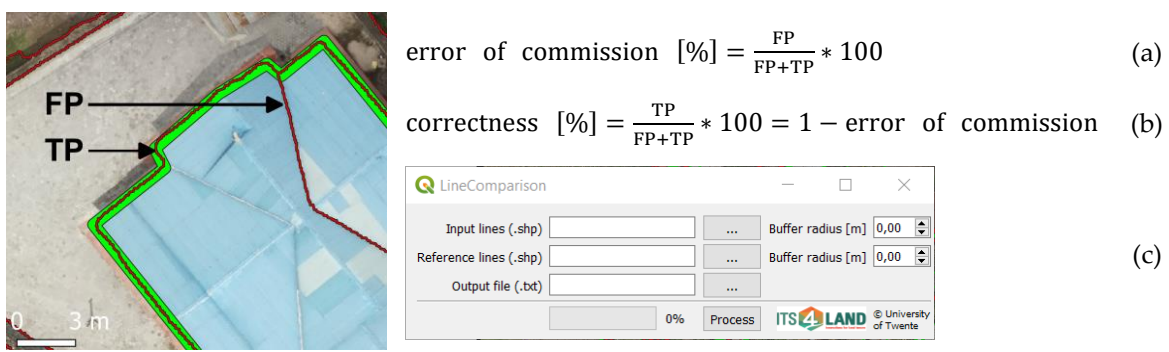

$$\text{error of commission } [\%] = \frac{\text{FP}}{\text{FP+TP}} * 100 \qquad (a)$$

$$\text{correctness } [\%] = \frac{\text{TP}}{\text{FP+TP}} * 100 = 1 - \text{error of commission} \qquad (b)$$

(c)

**Figure 4.** Spatial correctness based on overlaying the buffered delineation and reference data to compute pixels being True Positive (TP) or False Positive (FP). These pixels are then summated to calculate **(a)** the error commission and **(b)** the correctness. **(c)** We implemented the described procedure for line-based accuracy assessment in the 'LineComparison' QGIS plugin.

(iii) **Time Efficiency:** *How much time is saved by using the approach compared to manual delineation?*

Time efficiency investigates time savings of the delineation approach compared to manual delineation. Time savings can be obtained by reducing the amount of clicking and zooming required to delineate object outlines with the same completeness and correctness as for manual delineation. Time savings can also be obtained by reducing the processing time during the delineation process. Processing time of an automated workflow step that does not require interaction, should not be taken into consideration. In this study, only the time for the interactive delineation step is comparable to the time for manual delineation considering the same area and objects.

## 3. Results

As described in Section 2.2, the data was split into one tile of 250 x 250 m for training and two tiles of 250 x 250 m for testing for both study areas in Busogo and Mukingo. The dimensions resolution, input and location robustness, were investigated for building outlines, while application robustness, focused on cadastral boundaries. Results are shown in the following (Table 2, Figure 5) and are analyzed in Section 4. Detailed results can be found in the appendix (Table A2/A3).

For the building outlines, the training tile contained 1870 lines of which 30% were labeled as 'boundary' and 70% as 'not boundary' in Busogo. In this area, the testing tiles contained 4164 lines that covered 225 buildings in the reference data. In Mukingo, the training tile contained 967 lines of which 23% were labeled as 'boundary' and 77% as 'not boundary'. In this area, the testing tiles contained 1344 lines that covered 222 buildings in the reference data.

For the cadastral boundaries, the initial step of image segmentation required stronger over-segmentation, as the walls and fences to be extracted were less rich in contrast compared to building outlines. The training tile contained 5831 lines of which 19% were labeled as 'boundary' and 81% as 'not boundary' in Busogo. In Mukingo, the training tile contained 3558 lines of which 15% were labeled as 'boundary' and 85% as 'not boundary'. Visible parcel boundaries in the cadastral reference data delineated by two Rwandan surveying experts covered 37% of the study area in Busogo and 11% in Mukingo. In Mukingo, only a few objects demarcating cadastral boundaries were visible. These percentages align with the 22% of visible parcels observed in other rural Rwandan areas [58].

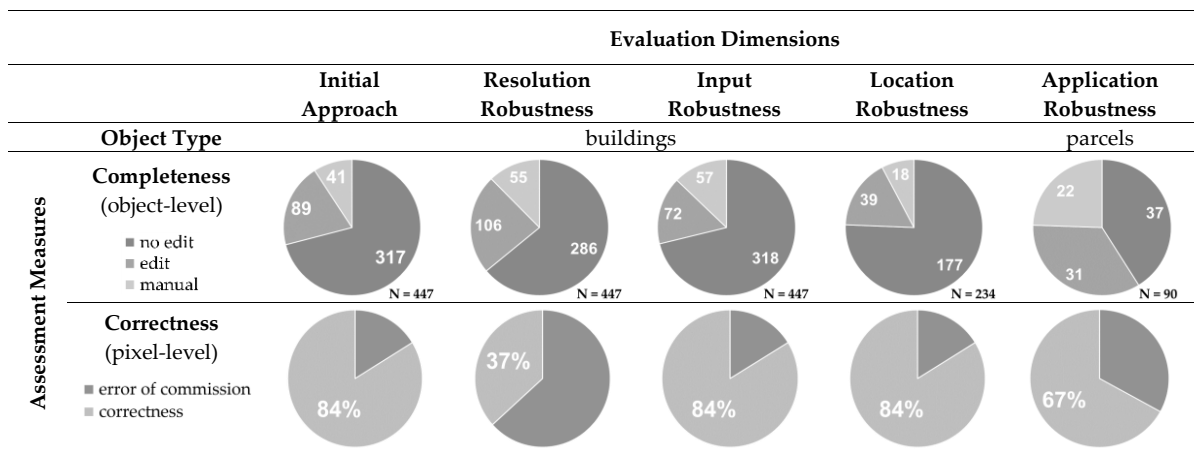

**Table 2.** Evaluation framework results.

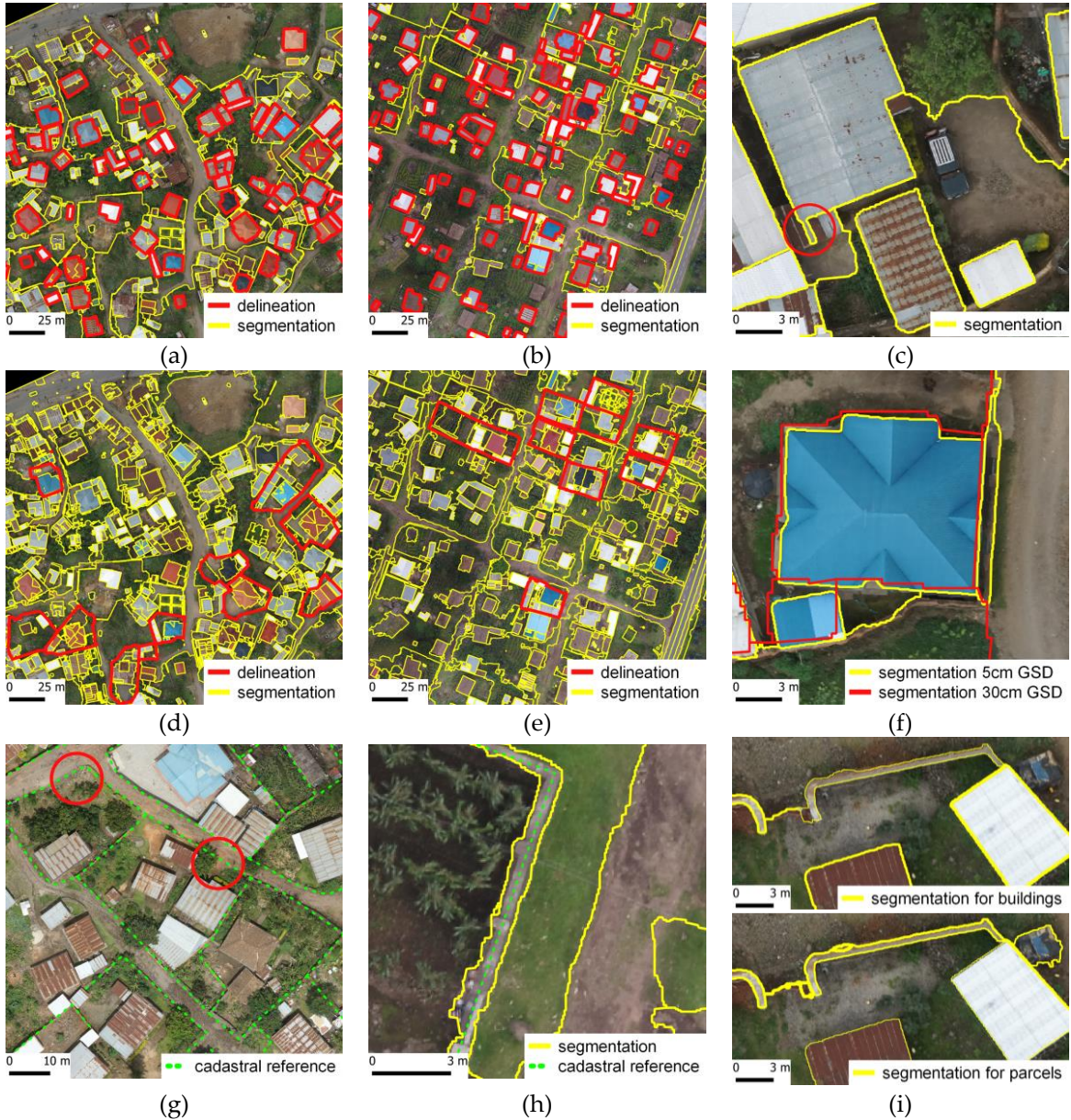

**Figure 5.** Examples of delineation results: (**a/b**) building delineation. (**d/e**) parcel delineation. (**c**) building segmentation requiring editing. (**f**) building segmentation from different resolutions. (**g**) visible parcels not demarcated by objects, but by context. (**h**) wall outline and centerline used for parcel delineation. (**i**) boundary classification trained to detect buildings (top) and parcels (bottom).

## 4. Discussion

In the following, results are discussed in general, followed by an analysis per dimension of the evaluation framework. As for buildings, 317 (71%) out of the 447 building objects can be directly extracted (Figure 5a/b), of which 84% are extracted with a correctness of 30 cm (Table 2). For the 86 (19%) buildings that needed editing, 82% required minor editing on <20% of the outline length, e.g., when the image segmentation did not provide a closed contour encompassing the entire building (Figure 5c). The 41 (9%) buildings that could not be delineated by the approach, were mostly missed entirely during image segmentation.

When comparing buildings and cadastral boundaries, buildings allow more unambiguous reference data. The data can be captured with precise labeling rules resulting in objects of a specific quantity with an accurate outline. For cadastral boundaries, defining precise labeling rules is more challenging: the reference data should capture visible object outlines that demarcate parcel boundaries. Which objects are considered as visible cadastral boundaries varies depending on the local context. The cadastral reference may contain parts that are not demarcated through visible objects, but can nevertheless be delineated based on context: if there is a gap between two walls delineating a parcel, the boundary is drawn as the shortest closing of this gap (Figure 5g). Similarly, the delineator might delineate the centerline of a wall as a cadastral boundary, while the boundary delineation approach captures the wall's in- or outline (Figure 5h).

(I) **Resolution Robustness:** *Can the approach be applied to data of a different resolution?*
The approach can be applied to data of a different resolution when accepting a lower spatial accuracy (Table 2). The completeness remains similar to the initial approach: from the data of 5cm GSD, 317 (71%) buildings were directly extracted, while this amount decreases slightly to 286 (64%) for data of 30 cm GSD. However, when buffering the reference data with 30cm, only 37% compared to 84% for the initial approach of the TP pixels fall within this buffer: the correctness on a pixel-level considering a 30cm buffer decreases. Thus, slightly fewer buildings are detected, but the detected outlines deviate more from the reference data (Figure 5f). To obtain the same correctness, i.e., 84%, as for the initial approach, the buffer would need to be increased to 80cm. In certain application cases, such as the delineation of agricultural fields, this lower spatial accuracy may be acceptable [54]. It can be concluded that depending on the application requirements, the approach is applicable to data of different resolutions that can be captured from UAVs, aircraft or satellite platforms. For cadastral mapping, that often requires high accuracies, UAVs can provide such high-resolution data, as well as the option to generate 3D objects and DSM information. Furthermore, they allow data capture in areas of limited accessibility, in which traditional direct surveying methods are not applicable [59]. The option to capture data in a fast, flexible and cost-effective manner, makes UAVs attractive in cases, where part of a cadastral map requires updating [60].

(II) **Input Robustness:** *Can the input data required for the approach be altered?*
The approach produces similar results regarding completeness and correctness when the DSM information is omitted. The number of buildings that needed editing in the initial approach slightly decreased by 4%. These 4% of buildings now require manual delineation. The results indicate that the buildings in the study areas are well distinguishable by color only: many buildings are separated by no or little space so that the separation can be done based on roof type differences and less on height differences. Furthermore, the segmentation produces little over-segmentation (Figure 5a/b), which reduces the need to sort out lines through the boundary likelihood generated in the boundary classification step. It can be assumed, that DSM information would be of higher support to distinguish lines in case of a stronger over-segmentation. In general, the approach is flexible in reducing or extending the amount of information used during the boundary classification step: features capturing information from additional data sources can be related to a line and subsequently be used by the RF classifier to produce a boundary likelihood.

(III) **Location Robustness:** *Can the approach be applied in a different location?*
The approach produces results similar to the initial approach regarding completeness and correctness when the classifier is trained on the Busogo study area and applied in the Mukingo study area, and vice versa (Table 2). The similarity in results compared to the initial approach can be explained by the little over-segmentation mentioned before: in case of a few lines available, the boundary likelihood plays a minor role in defining the optimal least-cost-path during the interactive delineation. Nevertheless, when comparing the range and distribution of boundary likelihood values, they overlap mainly with those of the initial approach. Thus, the classifier produces comparable boundary likelihoods when trained on the Busogo study area and applied in the Mukingo study area, and vice versa. This can be explained by similar characteristics of both peri-urban Rwandan scenes. By having a reusable classifier, the effort of applying the approach is reduced. The reusability depends on the similarity of objects to be delineated: if the classifier is trained to detect buildings in one scene, it can be reused to detect buildings of similar characteristics in a second scene. However, if the characteristics of the input data, the scene or the object change, retraining might be necessary. Compared to other state-of-the-art approaches that require large amounts of training data, retraining is of minor effort for this approach: only one tile of 250 x 250 m with manually labeled outlines was used for training. In general, the approach is not only transferable across similar scenes, but also across different objects: when trained to detect buildings, the boundary likelihood produced by the classifier is high for these objects. When trained to detect cadastral boundaries represented through walls and fences, building outlines get assigned a lower boundary likelihood (Figure 5i, the thinner the line, the lower the boundary likelihood). Such a transferability is beneficial for cadastral mapping that requires an object delineation not only across different scenes, but also across different objects.

(VI)     **Parameter Robustness:** *Does the approach depend on many or sensitive parameters?*
The approach requires minor parameter setting: only during image segmentation, one parameter can be altered. This parameter in range [0; 1] regulates over- and under-segmentation, i.e., the amount of extracted object outlines. For most cases, an over-segmentation is acceptable, since the subsequent steps of boundary classification and interactive delineation support filtering out irrelevant lines. Optimizing the parameter to only capture relevant lines can reduce the processing time and delineation effort in these subsequent steps. Under-segmentation that generates fewer lines, would still enable execution of the approach, but would reduce the approach's potential to be superior to manual delineation. Hence, as long as under-segmentation is avoided, settings of the segmentation parameter leading to different amounts of over-segmentation have only little effect on the quality of the final object outlines.

(IV)     **Application Robustness:** *Can the approach be applied to delineate different objects?*
The approach can be applied for the delineation of buildings and visible cadastral boundaries demarcated through walls and fences (Table 2, Figure 5d/e). The approach is most suited for areas in which a large portion of cadastral boundaries is visible. This is not entirely the case in the study areas, in which only 37% (Busogo) and 11% (Mukingo) of the area are covered by visible parcels. For the visible parcels, the approach can be used to delineate 37 of 90 (41%) parcels without further editing and 31 of 90 (34%) parcels that require minor editing (Table 2). The higher percentage of editing for parcels than for buildings can be explained by the outlines' complexity: parcel outlines run along different objects with varying characteristics, while building outlines are more consistent. Not all complexities are correctly captured during image segmentation and thus require editing.
The approach has now been investigated for numerous objects demarcating cadastral boundaries, such as roads [39], buildings, walls, and fences. Which objects demark visible cadastral boundaries is location-depended. Compared to manual delineation, which is the current procedure for indirect surveying [24], the approach obtains the highest time savings for areas in which boundaries are visible, long and curved. This has been shown for roads [39] and is assumed to hold true for agricultural field boundaries. The walls and fences investigated in this study are often covered by vegetation and not built consistently (Figure 5g). They are thus less suitable to derive representative time efficiency measures. Such boundaries are harder to extract during image segmentation so that the time efficiency for short and straight visible parts remains comparable to that for manual delineation. The approach could be improved by adding further functionalities: (i) allowing the user to add or move nodes and lines during delineation, (ii) automatically adjusting incorrect corners in the result (Figure 5c), and (iii) providing the option to apply the method without the boundary classification step as it adds little value for cases of clear and few boundaries detected through the image segmentation step (Figure 5a/b). Given the complexity of cadastral boundary characteristics, automating the delineation of these boundaries remains challenging. Despite considerable progress in object extraction and classification from remote sensing data, Höhle points out that a gap often remains between the result of an automatic approach and the desired map product [6]. Similarly, Chen et al. constitute a considerable amount of manual intervention required in most existing methods [35]. An alternative to our current approach may be a deep learning based approach: significant progress has been achieved in object extraction through deep learning [61-63]. For building extraction, deep learning has shown substantial improvements with high accuracies achieved mostly for buildings of consistent and regular roof sizes and shapes [35]. Applying similar approaches for boundary extraction is however not trivial: deep learning approaches require large amounts of training data, thus cadastral data and remote sensing data. Current governmental open data initiatives to publically share such data [64] and open-service data portals for aerial imagery [65] could contribute in generating sufficient training data and thus in developing deep learning approaches for an automated extraction of cadastral boundaries. In any case, the delineation cannot be fully automated at the current state, since the extracted outlines require (legal) adjudication and incorporation of local knowledge from human operators to create final cadastral boundaries.

## 5. Conclusion

This study contributes to recent advances of object extraction in remote sensing. The evaluation framework is applied to the proposed delineation approach: it investigates the approach's robustness (I) to use imagery of a different resolution, (II) to integrate DSMs as input, (III) to use a classifier trained on a different location, (IV) to extract different objects with altered parameter settings, (V) and to be applied for cadastral mapping. The evaluation framework can be reused during the development of future object delineation approaches in remote sensing.

The boundary delineation approach [39] is substantially improved and extended in this study. Compared to manual delineation the proposed approach is superior with regard to: (i) high spatial accuracy of boundaries that precisely follow object outlines, (ii) facilitated delineation procedure that merely requires selecting predefined nodes on a zoomed-out level, and (iii) encapsulation of delineator's knowledge during the boundary classification step. Nevertheless, it remains to be seen whether these aspects suffice to replace manual delineation currently used, e.g., in indirect surveying to delineate cadastral boundaries.

For cadastral mapping, it is important to note that, unlike described in previous studies [24,42], visible boundaries delineated in indirect surveying are not always demarcated by objects, but equally by their context. The development of automated approaches in cadastral mapping should thus not only rely on extracting objects, but also on closing boundaries based their context. In any case, extracted boundaries require (legal) adjudication and incorporation of local knowledge, limiting the scope of automated approaches. Therefore, we advise future studies to focus on the interactive part that bridges the gap between automatically generated results and the desired product, i.e., cadastral boundaries. Our future work will further investigate the interactive part: together with surveyors and stakeholders responsible for delineation tasks, the approach will be tested and analyzed to achieve a broader superiority over manual delineation.

**Funding:** This work was supported by the Horizon 2020 program of the European Union [project number 687828].

**Acknowledgments:** We are grateful to Claudia Stöcker for capturing and processing the UAV data, as well as to Emmanuel Nyandwi and Oscar Kanyamboneza for delineating the cadastral reference data.

**Author Contributions:** Sophie Crommelinck designed and conducted this study. She developed the object delineation approach and the evaluation framework. Mila Koeva, Michael Ying Yang and George Vosselman contributed to both developments and to the analysis of the results of this study. The manuscript was written by Sophie Crommelinck with contributions from Mila Koeva, Michael Ying Yang and George Vosselman.

**Declarations Interest:** None.

## Appendix

| Name | Description |
|------|-------------|
| **ID** | Unique number per line |
| **boundary** | Boundary label or likelihood in range [0; 1] |
| **vertices** | Number of vertices per line |
| **length [m]** | Length per line |
| **azimuth [°]** | Bearing in degrees between start and end of each line |
| **sinuosity** | Total line length divided by the shortest distance between start and end of each line |
| **red_grad** | Abs. difference between median of all red values lying within a 0.4 m buffer right and left of each line |
| **green_grad** | Same as red_grad for green of RGB |
| **blue_grad** | Same as red_grad for blue of RGB |
| **dsm_grad** | Same as red_grad for DSM |

**Table A1.** Features calculated per line to be used by the Random Forest (RF) classifier for boundary classification. The first two features are not used for the classification.

| extent | GSD [cm] | DSM | transferred | Mukingo error of commission [%] | Busogo error of commission [%] | Mukingo buildings no edit (N) | Busogo buildings no edit (N) | Mukingo buildings edit (N) | Busogo buildings edit (N) | Mukingo buildings manual (N) | Busogo buildings manual (N) |
|--------|----------|-----|-------------|------|------|------|------|------|------|------|------|
| 1 | 5 | yes | no | 16 | 17 | 103 | 72 | 17 | 39 | 11 | 11 |
| 2 | 5 | yes | no | 16 | 17 | 68 | 74 | 7 | 26 | 16 | 3 |
| 1 | 30 | yes | no | 66 | 60 | 83 | 69 | 27 | 41 | 21 | 12 |
| 2 | 30 | yes | no | 63 | 63 | 68 | 66 | 7 | 31 | 16 | 6 |
| 1/2 | 5 | yes | yes | 16 | 16 | 104 | 73 | 14 | 25 | 13 | 5 |
| 1 | 5 | no | no | 16 | 16 | 69 | 72 | 6 | 28 | 16 | 22 |
| 2 | 5 | no | no | 17 | 17 | 104 | 73 | 14 | 24 | 13 | 6 |

**Table A2.** Detailed results for building delineation.

| extent | GSD [cm] | DSM | transferred | Mukingo error of commission [%] | Busogo error of commission [%] | Mukingo boundaries no edit (N) | Busogo boundaries no edit (N) | Mukingo boundaries edit (N) | Busogo boundaries edit (N) | Mukingo boundaries manual (N) | Busogo boundaries manual (N) |
|--------|----------|-----|-------------|------|------|------|------|------|------|------|------|
| 1 | 5 | yes | no | 22 | 35 | 9 | 14 | 4 | 15 | 4 | 13 |
| 2 | 5 | yes | no | 37 | 36 | 3 | 11 | 0 | 12 | 0 | 5 |

**Table A3.** Detailed results for cadastral boundary delineation.